%% file: camera_ready.tex
\documentclass[10pt,twocolumn,letterpaper]{article}

\usepackage{cvpr}
\usepackage{times}
\usepackage{epsfig}
\usepackage{graphicx}
\usepackage{amsmath}
\usepackage{amssymb}
\usepackage{booktabs}
\usepackage{multirow}
\usepackage{caption}
\usepackage{xfrac}

\usepackage[pagebackref=true,breaklinks=true,letterpaper=true,colorlinks,bookmarks=false]{hyperref}

\cvprfinalcopy 

\def\cvprPaperID{1135} 

\ifcvprfinal\fi
\begin{document}
\title{Example-Guided Style-Consistent Image Synthesis from Semantic Labeling}
\author{Miao Wang$^{1}$ \qquad  Guo-Ye Yang$^{2}$ \qquad  Ruilong Li$^2$ \qquad  Run-Ze Liang$^2$ \\ 
Song-Hai Zhang$^2$  \qquad  Peter M. Hall$^3$ \qquad Shi-Min Hu$^{2,1}$\vspace{5pt}\\
$^1$ State Key Laboratory of Virtual Reality Technology and Systems, Beihang University\\
$^2$ Department of Computer Science and Technology, Tsinghua University, Beijing\\
$^3$ University of Bath
}

\twocolumn[{%
\renewcommand\twocolumn[1][]{#1}%
\maketitle

\begin{center}
\vspace{-15pt}
    \centering
    \includegraphics[width=0.9\textwidth]{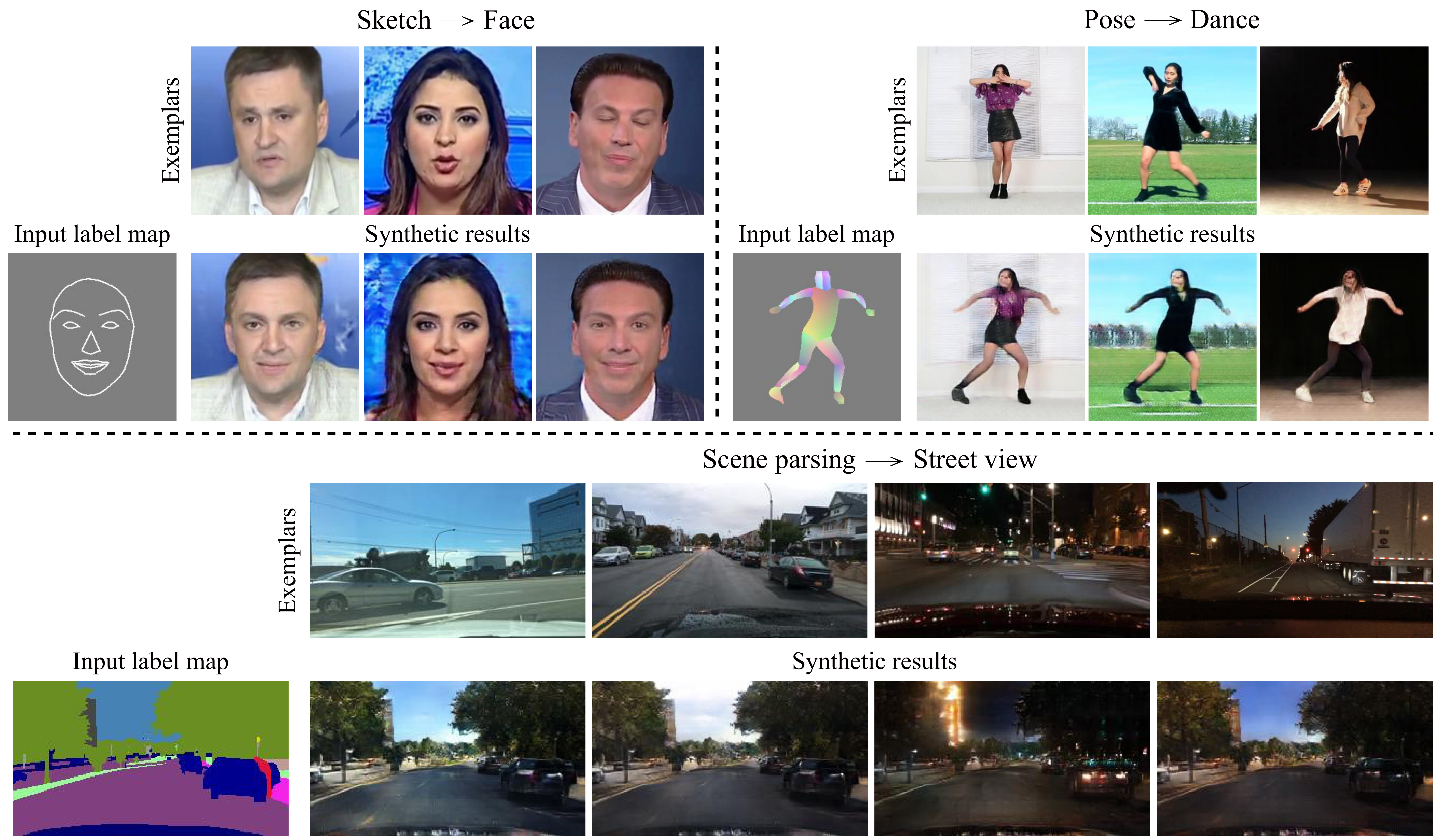}
    \vspace{-5pt}
    \captionof{figure}{We present a generative adversarial framework for synthesizing images from semantic label maps as well as image exemplars. Our synthetic results are photorealistic, semantically consistent to the label maps (facial expression, pose or scene segmentation map) and style-consistent with corresponding exemplars.}
    \label{fig:teaser1}
\end{center}%
}]


\begin{abstract}
\vspace{-6pt}
Example-guided image synthesis aims to synthesize an image from a semantic label map and an exemplary image indicating style.  We use the term ``style" in this problem to refer to implicit characteristics of images, for example: in portraits  ``style"  includes gender, racial identity, age, hairstyle; in full body pictures it includes clothing; in street scenes it refers to weather and time of day and such like. A semantic label map in these cases indicates facial expression, full body pose, or scene segmentation. 
We propose a solution to the example-guided image synthesis problem using conditional generative adversarial networks with style consistency.
Our key contributions are (i) a novel style consistency discriminator to determine whether a pair of images are consistent in style; (ii) an adaptive semantic consistency loss; and (iii) a training data sampling strategy, for synthesizing style-consistent results to the exemplar.
We demonstrate the efficiency of our method on face, dance and street view synthesis tasks. Code and data are available via: \href{https://github.com/cxjyxxme/pix2pixSC}{https://github.com/cxjyxxme/pix2pixSC}
    
\end{abstract}
\vspace{-15pt}
\section{Introduction}

In image-to-image translation tasks, mappings between two visual domains are learnt.
Various computer vision and graphics problems are addressed and formulated using the image-to-image translation framework, including super-resolution \cite{Ledig2016a,LapSRN}, colorization \cite{Gustav2016,zhang2016colorful}, inpainting \cite{pathakCVPR16context,IizukaSIGGRAPH2017}, style transfer \cite{Johnson2016Perceptual,luan2017deep} and photorealistic image synthesis \cite{pix2pix2016,chen2017photographic,pix2pixHD}. In the photorealistic image synthesis  problem, images are generated from abstract semantic label maps such as pixel-wise segmentation maps or sparse landmarks. In this paper, we study the problem of example-guided image synthesis. Given an input semantic label map $x$ and a guidance image $I$, the goal is to synthesize a photorealistic image, $y$, which is semantically consistent with the label map $x$, while being style-consistent with the exemplar $I$, so $(x,I) \mapsto y$. Style consistency is automatically determined: in portraits, style consistency refers to the fact that we want our synthetic output to be plausibly of the same genetic type as an input exemplar; in full body images style consistency means the same clothing; and in street scenes it includes such things as the same weather and time of day. Representative applications are shown in Figure \ref{fig:teaser1}.


Example-based image synthesis cannot be solved with a straightforward combination of photorealistic image synthesis based on pix2pixHD \cite{pix2pix2016,pix2pixHD} and style transfer \cite{luan2017deep}; the style of the input exemplar is not well kept in the synthetic result, see Figure~\ref{fig:compare_face_all}. Recently, example-guided image-to-image translation frameworks \cite{huang2018munit,DRIT,bao2018towards} are proposed using a disentangled model to represent content and style or identity and attributes, however they fail to synthesize photorealistic results from abstract semantic label maps. The challenges are multi-fold: first, the ground truth photorealistic result for each label map given an arbitrary exemplar is not available for training; second, the synthetic results should be photorealistic while semantically consistent with the source label maps; last but not least, the synthetic result should be stylistically consistent with the corresponding image exemplar. 

We present a method for this example-guided image synthesis problem with conditional generative adversarial networks.
We build on  the recent  pix2pixHD \cite{pix2pixHD} for image synthesis to ensure photorealism, with the crucial contributions of: 
\begin{itemize}
\item a novel style consistency discriminator to enforce style consistency of a pair of images (see Section~\ref{subsubsec:styleloss})
;
\vspace*{-6pt}
\item an adaptive semantic consistency loss to ensure quality  (see Section~\ref{subsubsec:labelloss});
\vspace*{-6pt}
\item a data sampling strategy that ensures we need only a weakly supervised approach for training (see  Section~\ref{sec:sample}).
\end{itemize}



\input{RelatedWork}
\input{ImageSynthesis_Peter2}

\input{Results_Peter}

\vspace{-5pt}
\section{Conclusions}
\vspace{-5pt}
In this paper, we present a novel method for example-guided image synthesis with style-consistency from general-form semantic labels. During network training, we propose to sample style-consistent and style-inconsistent image pairs from video to provide style awareness to the model. Beyond that, we introduce the style consistency adversarial losses and the style consistency discriminator, as well as the semantic consistency loss with adaptive weights, to produce plausible results. Qualitative and
quantitative results in different applications show that the proposed model produces realistic and style-consistent images better than those from prior arts.

\textbf{Limitations and Future Work.} Our network is mainly trained on cropped video data whose resolution is limited (\emph{e.g.} $256\times 256$), we did not use the multi-scale architecture as pix2pixHD did for high-resolution image synthesis (\emph{e.g.} $512\times 512$ resolution or more).
Moreover, the synthetic background in face and dance image synthesis tasks may be blurry, because the semantic labels do not specify any background scenes. Lastly, we have demonstrated the efficiency of our method in several synthesis applications, however the results in other applications could be effected by the performance of the state-of-the-art semantic labeling function $F(\cdot)$.  In the future, we plan to extend this framework to video domain \cite{wang2018vid2vid} and synthesize style-consistent videos to given exemplars.

\textbf{Acknowledgements.} We thank the anonymous reviewers for the valuable discussions. This work was supported by the Natural Science Foundation of China (Project Number: 61521002, 61561146393). Shi-Min Hu is the corresponding author.

{\small
\bibliographystyle{ieee_fullname}
\bibliography{egbib}
}

\appendix
\section{Datasets}

As described in the main manuscript, we evaluate our model on face, dance and street view image synthesis tasks, using following datasets and semantic functions:

\textbf{Sketch$\rightarrow$Face.} We use the real videos in the FaceForensics dataset \cite{roessler2018faceforensics}, which contains $854$ videos of reporters broadcasting news. We use the image sampling strategy described in Section \ref{sec:sample} of the main manuscript to acquire training image pairs from video, then apply face alignment algorithm \cite{king2009dlib} to localize facial landmarks, crop facial regions and resize them to size $256\times256$. We sample $20,000$ images from  videos for training and $500$ images from distinct videos for testing. The detected facial landmarks are connected to create face sketches; this is the function $F(\cdot)$, in both training set and test set. For each sketch extracted from a training image, we randomly sample $30$ guidance images from other videos for training, and for each testing sketch, we randomly sample $5$ guidance images from other videos for testing.

\textbf{SceneParsing$\rightarrow$StreetView.}  We use the BDD100k dataset \cite{yu2018bdd100k} to synthesize street view images from pixel-wise semantic labels ({\em i.e.} scene parsing) maps. For each street view image in the dataset, the corresponding scene parsing map and \textsc{weather} and \textsc {timeofday} attributes are provided. Based on these attributes, we divide images into $13$ style groups as listed in Table \ref{tab:bdd100k}, then sample style-consistent image pairs inside each group and style-inconsistent image pairs between groups.
The training set contains $2,000$ images and test set contains $400$ images, both resized to width $256$. We use scene parsing network DANet \cite{fu2018dual} as the function $F(\cdot)$ for each street view image during testing. For each scene parsing map, we randomly select an image inside each style group as the guidance, both in training and testing phases.

\begin{table}[!t]
\centering

\begin{tabular}{c | c c}

 Group & Weather & Timeofday \\
\hline
1 & - & Night \\
2 & Foggy & Dawn or Dusk \\
3 & Overcast & Daytime \\
4 & Rainy & Dawn or Dusk \\
5 & Snowy & Dawn or Dusk \\
6 & Clear & Dawn or Dusk \\
7 & Foggy & Daytime \\
8 & Partly cloudy & Dawn or Dusk \\
9 & Rainy & Daytime \\
10 & Snowy & Daytime \\
11 & Clear & Daytime \\
12 & Overcast & Dawn or Dusk \\
13 & Partly cloudy & Daytime \\

\end{tabular}
\caption{Style groups we used to categorize BDD100K street view images.}
\label{tab:bdd100k}
\end{table}

\textbf{Pose$\rightarrow$Dance.} We downloaded $150$ solo dance videos from YouTube, cropped out the central body regions and resized them to size $256\times256$. As the number of videos is small, we evenly split each video into the first part and the second part along the timeline, then sample training data only from the first parts and sample testing data only from the second parts of all the videos. The function $F(\cdot)$ is implemented using concatenated pre-trained DensePose \cite{Guler2018DensePose} and OpenPose \cite{cao2017realtime} pose detection results to provide pose labels. As a result, we have $35,000$ images  for training and $500$ images for testing. For each pose extracted from a training image, we randomly sample $30$ guidance images from other dancing videos, and for each testing pose, we randomly sample $5$ guidance images from other dancing videos.

\section{Network Architectures}


\subsection{Generator}
We follow the naming convention used in Johnson et al. \cite{Johnson2016Perceptual}, CycleGAN \cite{CycleGAN2017} and pix2pixHD \cite{pix2pixHD}. Let \texttt{c7s1-k} denote a $7\times 7$ Convolution-InstanceNorm-ReLU layer with $k$ filters and stride 1. \texttt{dk} denotes a $3\times 3$ Convolution-InstanceNorm-ReLU layer with $k$ filters and stride $2$. Reflection padding is used to reduce boundary artifacts. \texttt{Rk$\times$t} denotes residual blocks each contains two $3\times 3$ convolutional layers with $k$ filters, repeated $t$ times. \texttt{uk} denotes a $3\times 3$ fractional-strided-Convolution-InstanceNorm-ReLU layer with $k$ filters and stride $0.5$.

The architecture of generator is represented as:\\

\texttt{c7s1-64, d128, d256, d512, d1024, R1024$\times$9, u512, u256, u128, u64, c7s1-3}

\subsection{Discriminators}

We use $35\times 35$ PatchGAN \cite{pix2pix2016} in both of the two discriminators $D_R$ and $D_{SC}$. Let \texttt{Ck} denote a $4\times 4$ Convolution-InstanceNorm-LeakyRU layer with $k$ filters and stride $2$. The last layer is send to an extra convolution layer to produce a $1$ dimensional output. InstanceNorm is not used for the first \texttt{C64} layer. Leaky ReLU slope is set as $0.2$.

The architectures of $D_R$ and $D_{SC}$ are identical:\\

\texttt{C64, C128, C256, C512}

\section{Training Details}

All the networks were trained from scratch. Weights were initialized from a Gaussian distribution with mean 0 and standard deviation 0.02. In the first 250K iterations, the learning rate was fixed as 0.0002 with the adversarial style-consistency loss $\mathcal{L}_{\text{SCAdv}}$ turned-off. In the next 250K iterations, we turned on the $\mathcal{L}_{\text{SCAdv}}$ loss. In the final 500K iterations, the learning rate linearly decayed to zero with all the losses turned-on.

The models were trained on an NVIDIA TITAN 1080 Ti GPU with 11GB memory. The inference time is about 8-10 milliseconds per $256\times256$ image.

\section{Additional Results}

\begin{figure}[t]
\begin{center}
\includegraphics[width=\linewidth]{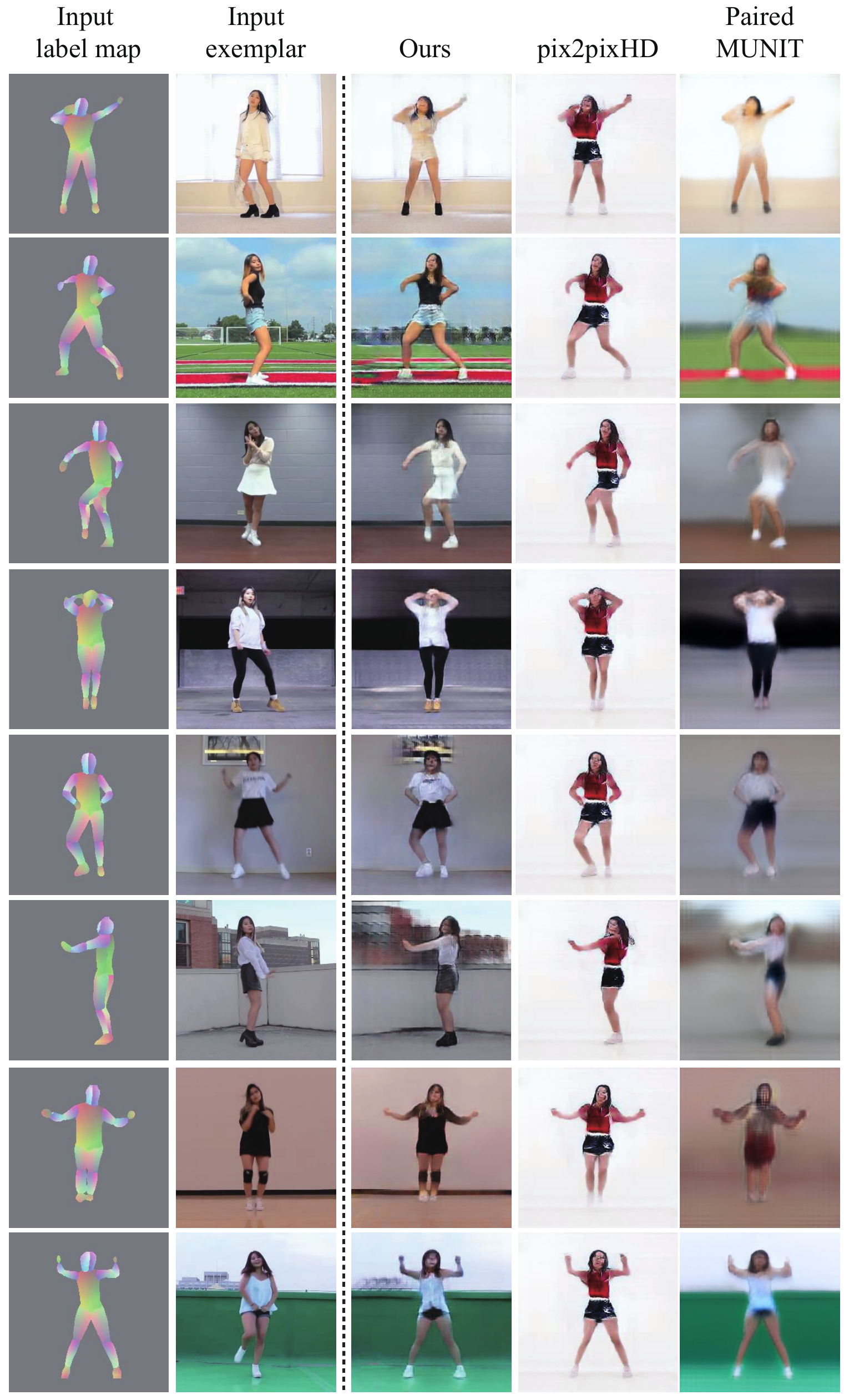}
\end{center}
\vspace{-18pt}
   \caption{Example-based dance image synthesis YouTube Dance dataset. The first column shows the input pose labels, the second column shows the input style examples, next columns show the results from our method, pix2pixHD and PairedMUNIT.}
\label{fig:compare_dance}
\vspace{-10pt}
\end{figure}

In Figure \ref{fig:compare_dance} and following pages, we show further experimental results from our method and baselines.

\begin{figure*}[t]
\begin{center}
\includegraphics[width=\linewidth]{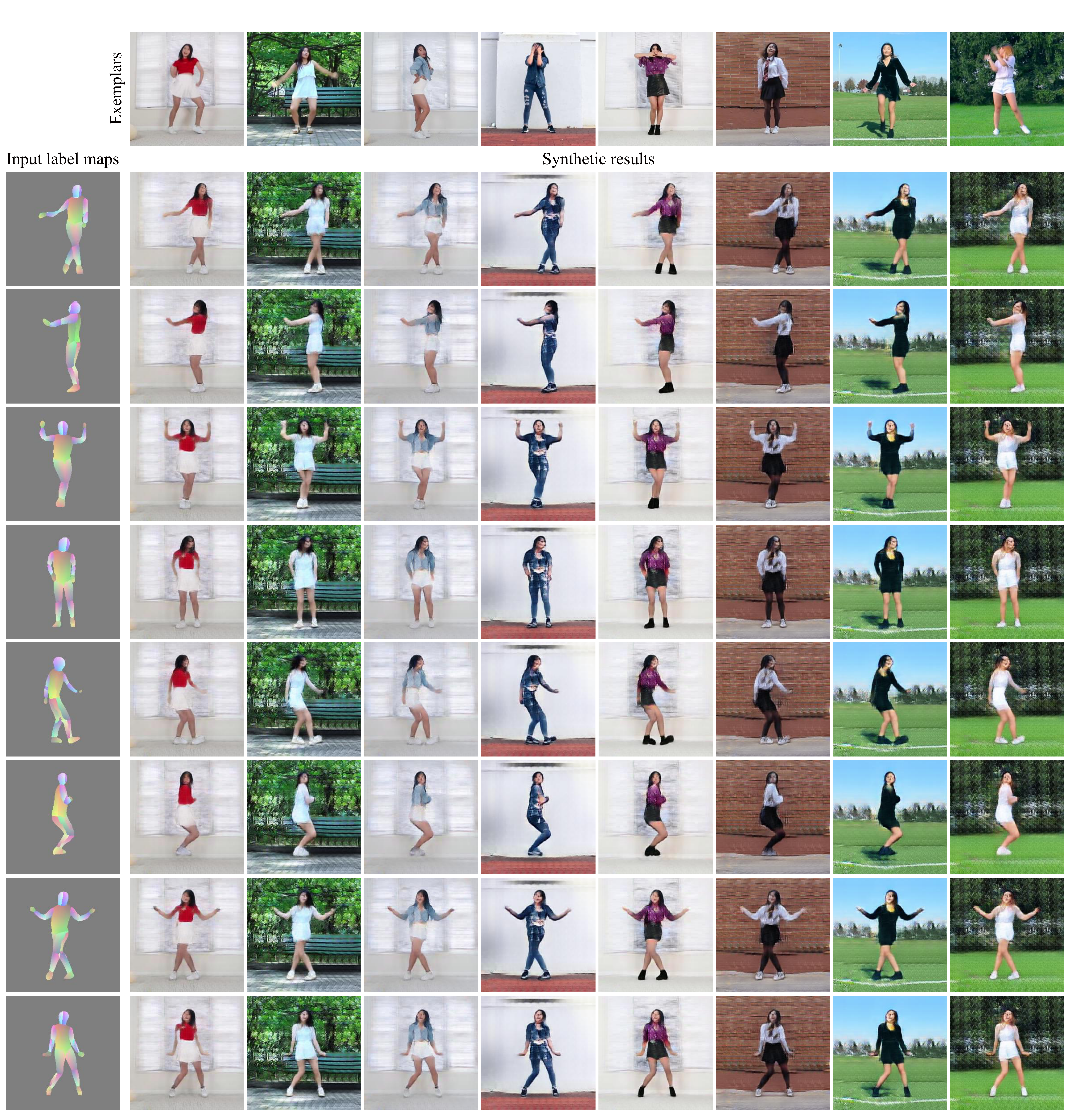}
\end{center}
\vspace{-18pt}
   \caption{More results of dance synthesis. The first column shows input pose maps. The first row shows input dance exemplars. Other images are the synthetic dance results.}
\label{fig:compare_face_all}
\vspace{-10pt}
\end{figure*}

\begin{figure*}[t]
\begin{center}
\includegraphics[width=\linewidth]{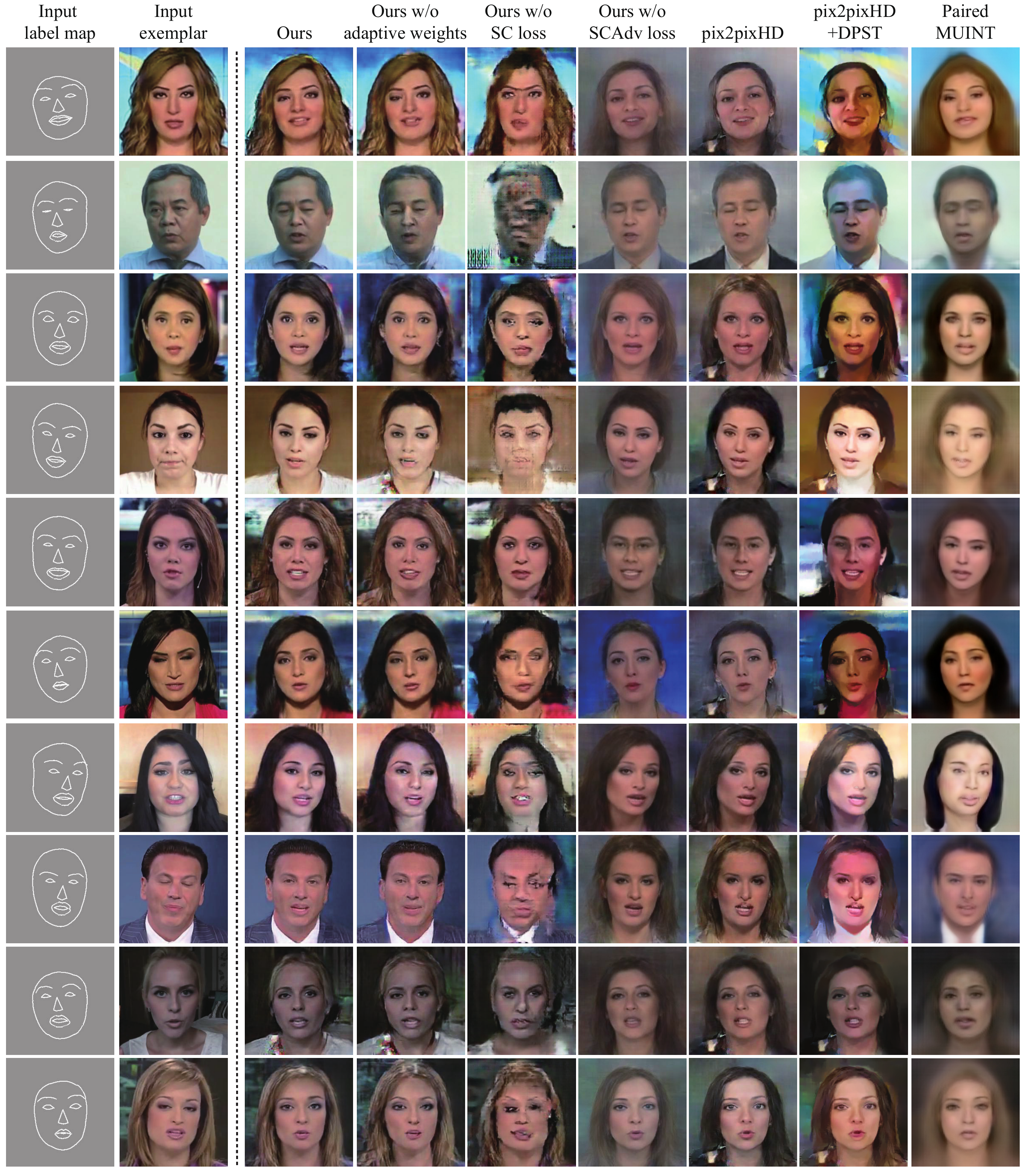}
\end{center}
\vspace{-18pt}
   \caption{Example-based face image synthesis on the FaceForensics dataset. The first column shows the input labels, the second column shows the input style example, next columns show the results from our method and our ablation studies, pix2pixHD, pix2pixHD+DPST and PairedMUNIT.}
\label{fig:compare_face_all}
\vspace{-10pt}
\end{figure*}

\begin{figure*}[t]
\begin{center}
\includegraphics[width=\linewidth]{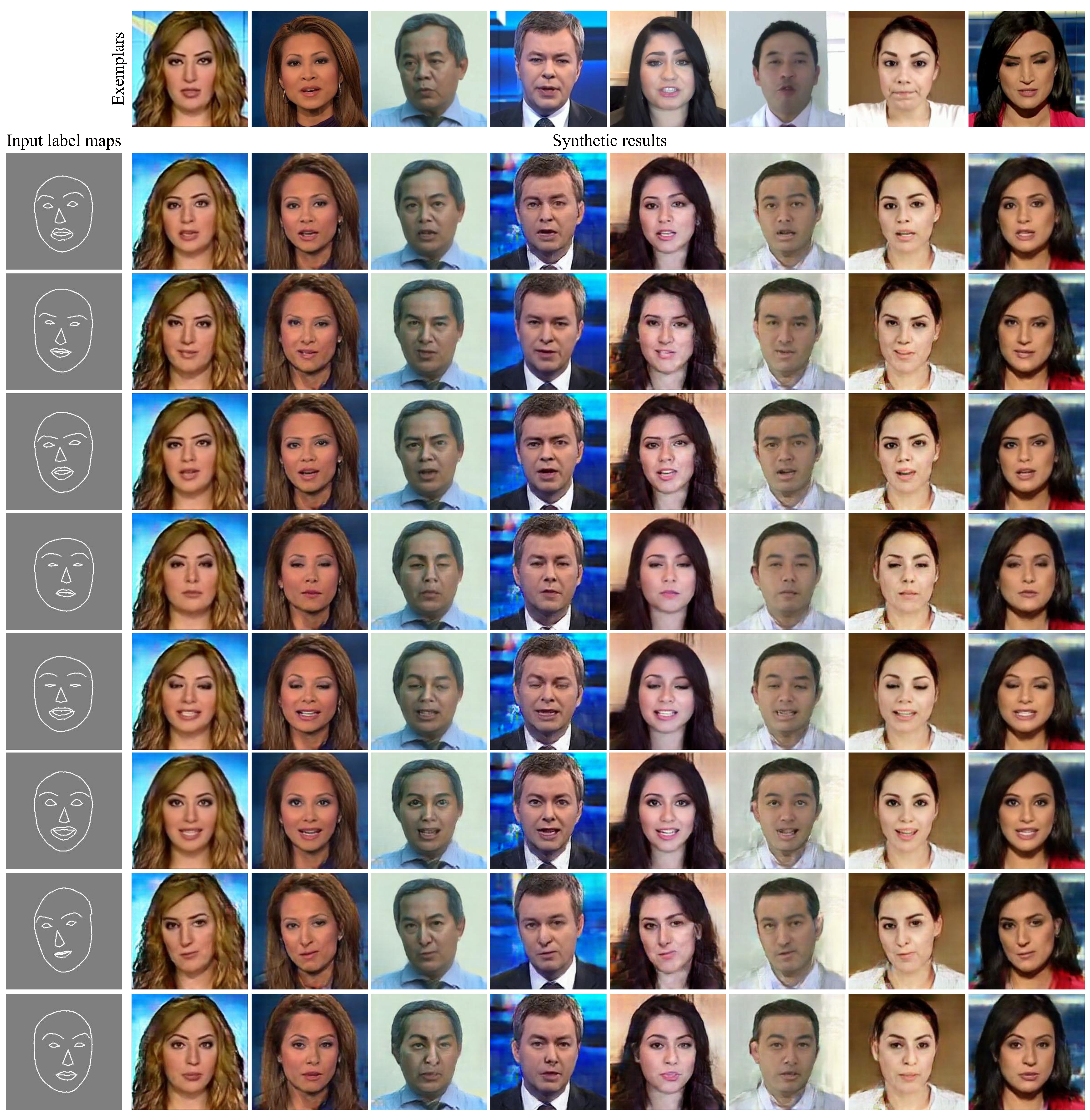}
\end{center}
\vspace{-18pt}
   \caption{More results of face synthesis. The first column shows input sketch maps. The first row shows input face exemplars. Other images are the synthetic face results.}
\label{fig:compare_face_all}
\vspace{-10pt}
\end{figure*}

\begin{figure*}[t]
\begin{center}
\includegraphics[width=\linewidth]{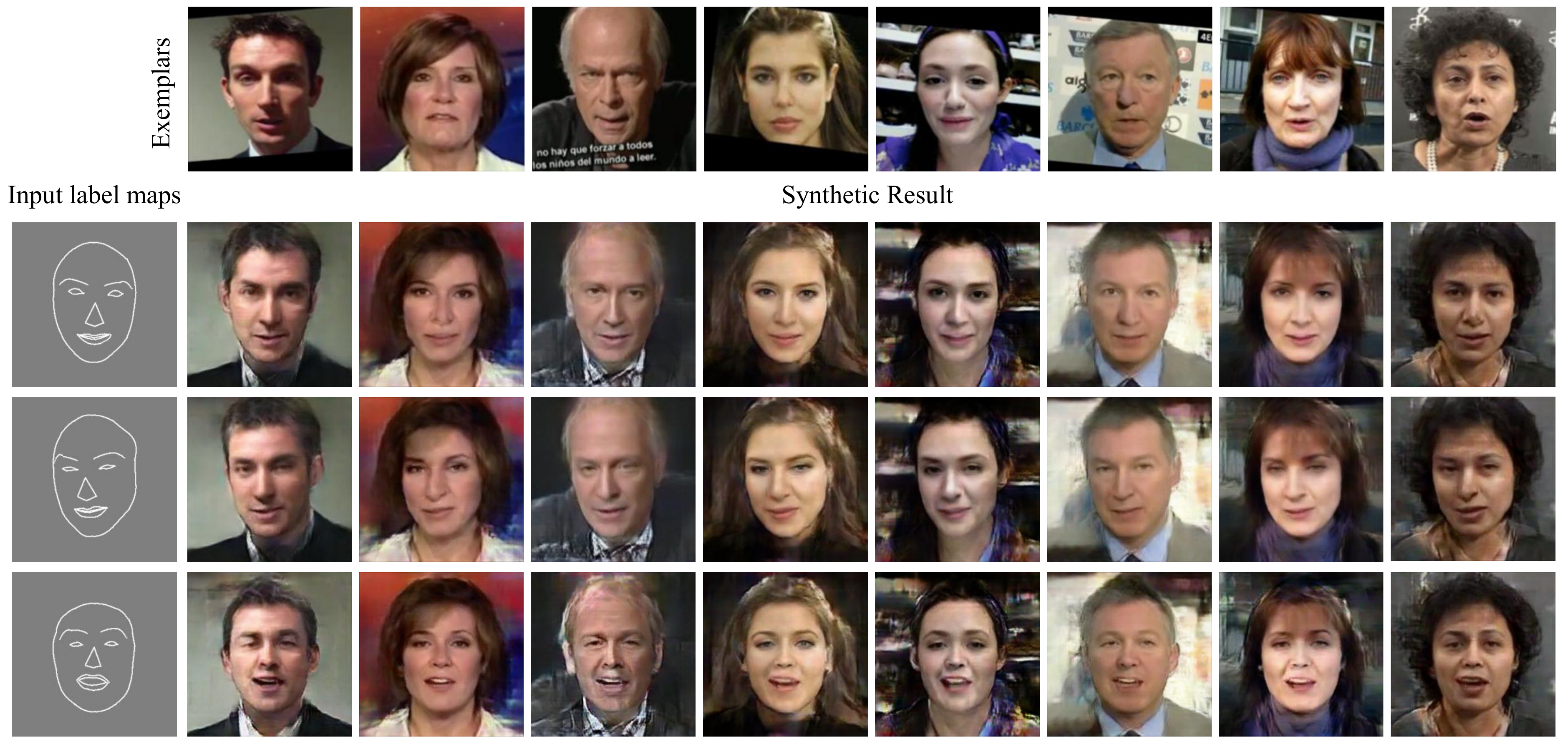}
\end{center}
\vspace{-18pt}
   \caption{More in-the-wild Sketch$\rightarrow$Face results. The model is trained on our training dataset and tested on Internet images.}
\label{fig:inthewild}
\vspace{-10pt}
\end{figure*}

\begin{figure*}[t]
\begin{center}
\includegraphics[width=0.8\linewidth]{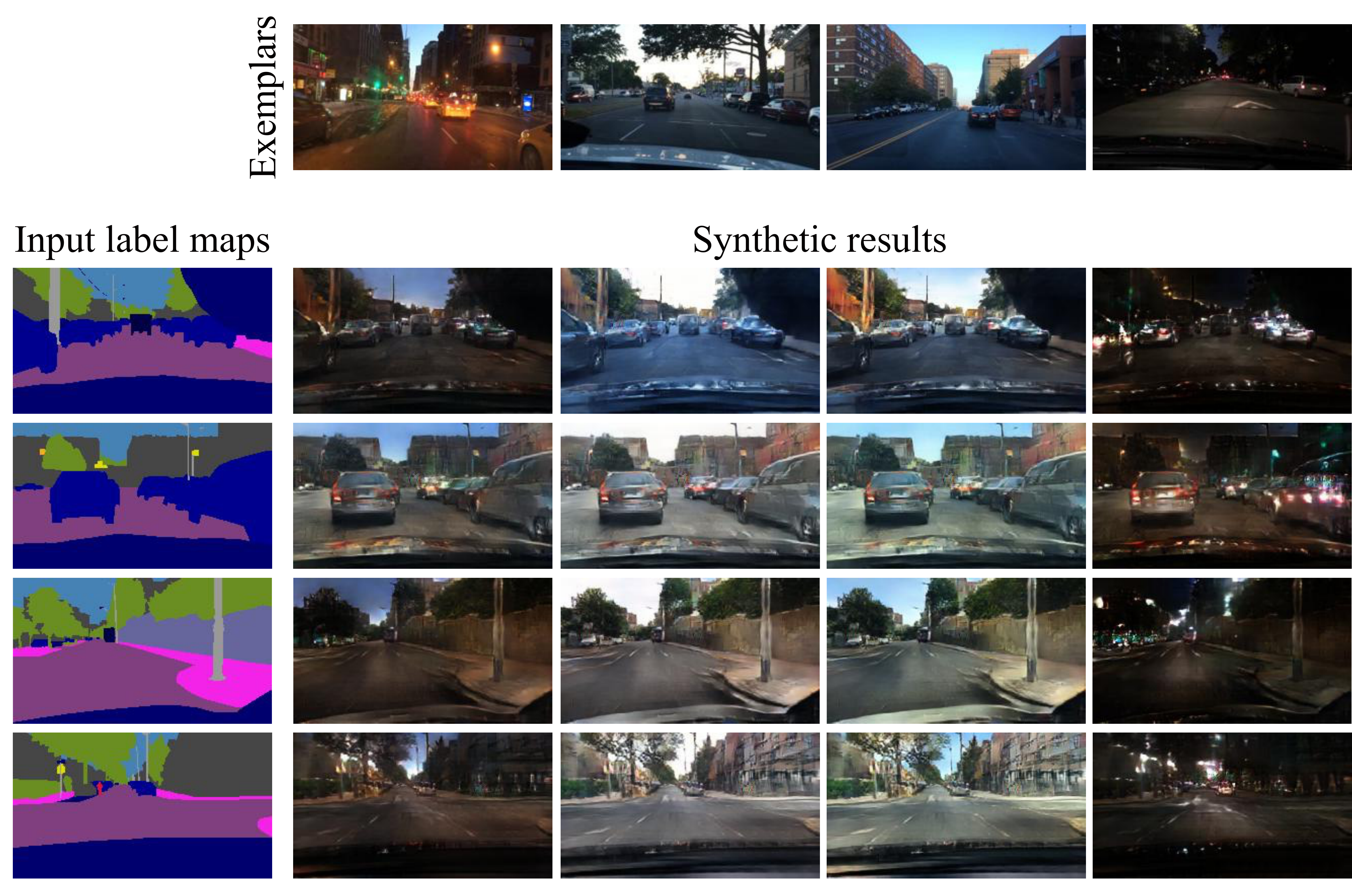}
\end{center}
\vspace{-18pt}
   \caption{More results of street view synthesis. The first column shows input segmentation maps. The first row shows input exemplars. Other images are the synthetic street view results.}
\label{fig:compare_street}
\vspace{-10pt}
\end{figure*}
\end{document}

%% file: RelatedWork.tex
\section{Related Work}

\paragraph{Generative Adversarial Networks.} In recent years, generative adversarial networks (GANs) \cite{goodfellow2014generative,arjovsky2017wasserstein} for image generation have progressed rapidly \cite{pix2pix2016,pix2pixHD}. Driven by adversarial losses, generators and discriminators compete with each other: discriminators aim to distinguish the generated fake images from the target domain; generators try to fool discriminators. Technologies to improve GANs include: progressive GANs \cite{huang2017sgan,zhang2017stackgan,karras2017progressive}, training objective and process designs \cite{salimans2016improved,arjovsky2017wasserstein,mao2017least,shrivastava2017learning}, {\em etc}. In this paper, we use GANs for example-guided image generation with style consistency awareness.

\vspace{-10pt}
\paragraph{Image-to-Image Translation and Photorealistic Image Synthesis.} The goal of image-to-image translation is to translate images from a source domain to a target domain. Isola \etal \cite{pix2pix2016} proposed the conditional GAN framework for various image-to-image translation tasks with paired images for supervision. Wang \etal \cite{pix2pixHD} extended this work for high-resolution image synthesis and interactive manipulation. 
Recently, researchers proposed to solve the unsupervised image-to-image translation problem with cycle consistency to overcome the lack of unpaired training data \cite{CycleGAN2017,kim2017learning,liu2017unsupervised,zhu2017toward,huang2018munit,DRIT,chang2018pairedcyclegan}. Photorealistic image synthesis \cite{chen2017photographic,qi2018semi,pix2pixHD} is a specific application of image-to-image translation, where images are synthesized semantically from abstract label maps. Chen \etal \cite{chen2017photographic} proposed a cascade framework to synthesis high-resolution images from pixel-wise labeling maps. Wang \etal \cite{pix2pixHD} proposed a framework for instance-level image synthesis with conditional GANs. 

Very recently, a few works \cite{pmlr-v80-hoffman18a,huang2018munit,DRIT,liqianExample} have been proposed to transfer the style or attributes of an exemplar to the source image, where the images belong to photorealistic domains (\emph{aka} domain adaptation). Our goal differs from these works by aiming at synthesizing photos from an abstract semantic label domain rather than a photorealistic image domain. Zheng \etal \cite{Zheng2017} proposed a clothes changing system to change the clothing of a person in image. Chan \etal \cite{chan2018everybody} presented a network to synthesize a dance video from a target dance video and an source exemplar video. Different from our model, it was trained for every input exemplar video. Ma \etal \cite{DBLP:conf/nips/MaJSSTG17} proposed to synthesize person images from pose keypoints. We show in Section \ref{sec:experiments} that our method outperforms the state-of-the-art methods. 

\vspace{-10pt}
\paragraph{Style Transfer.} Style transfer is a long-standing problem in computer vision and graphics, which aims to transfer the style of a source image to a target image or target domain. Some approaches \cite{Hertzmann:2001:IA:383259.383295,gatys2016image,Johnson2016Perceptual,luan2017deep,huang2017arbitrary,Liao:2017:VAT:3072959.3073683,gu2018arbitrary,chang2018pairedcyclegan,Huang_2017_CVPR} transfer style based on single exemplar, where others learn a general style of a target domain with a holistic sense \cite{CycleGAN2017,huang2018munit,DRIT,chen2018cartoongan}. Similar to our model, the PairedCycleGAN model \cite{chang2018pairedcyclegan} uses a style discriminator to distinguish whether a pair of facial images wear the same make-up in the making-up application. However, in their discriminator, the input image pair must be accurately aligned  via warping;  a generator is learned for each facial component. Our style consistency discriminator, in contrast, provides a general solution for image synthesis from both sparse labels (\emph{e.g.} sketch and pose) and pixel-wise dense labels (\emph{e.g.} scene parsing).


%% file: ImageSynthesis_Peter2.tex
\section{Example-guided Image Synthesis}

In this section, we first review the baseline model pix2pixHD \cite{pix2pixHD}, then describe our method, a conditional generative adversarial network for synthesizing photorealistic images from semantic label maps given specific exemplars. Finally we show how to appropriately prepare training data for our framework. 

\subsection{The pix2pixHD Baseline}

The pix2pixHD \cite{pix2pixHD} is a powerful image synthesis and interactive manipulation framework based on the pioneering conditional image-to-image translation method pix2pix \cite{pix2pix2016}. Let $x$ be a label map from a semantic label domain $\mathcal{X}$, the goal of pix2pixHD is to synthesize an image $y$, from $x$: $x \mapsto y$. It consists of a hierarchically integrated generator $G$ and multi-scale discriminators $\mathcal{D}=\{D_1,D_2,D_3\}$ to handle high-resolution synthesis tasks. The goal of the generator $G$ is to translate semantic label maps to photorealistic images, and the objective of the discriminators is to distinguish generated fake images from real ones at different resolution. The training dataset $\{(x_i,y_i)\}$ consists of pairs of label map $x_i$ and corresponding real image $y_i$.

The pix2pixHD optimizes a multi-task problem with a standard GAN loss $\mathcal{L}_{\text{GAN}}$  and  feature matching loss $\mathcal{L}_{\text{FM}}$:
\vspace{-15pt}

\begin{equation}
\min_G{(\max_{\mathcal{D}}{\sum_{k=1,2,3}\mathcal{L}_{\text{GAN}}(G,D_k)+\lambda\mathcal{L}_{\text{FM}}(G,D_k)})},
\end{equation}
where $\mathcal{L}_{\text{GAN}}(G,D_k)$ is the standard GAN loss given by:
\vspace{-5pt}
\begin{equation}
   \mathbb{E}_{(x,y)}[\log D_k(x,y)] + \mathbb{E}_{x}[\log(1-D_k(x, G(x)))],
\end{equation}
and $\mathcal{L}_{\text{FM}}(G,D_k)$ is the feature matching loss given by:
\vspace{-5pt}
\begin{equation}
   \mathbb{E}_{(x,y)}\sum_{i=1}^T{\frac{1}{N_i}[||D^i_k(x,y)-D^i_k(x,G(x))||_1]},
\end{equation}
where $T$ is the layer size and $N_i$ is the feature size in corresponding discriminator layer. An optional perceptual loss is introduced as the $L_1$ loss between pre-trained VGG network \cite{simonyan2014very} features.

One appealing feature of pix2pixHD is the instance-level image manipulation with a feature embedding technique. Given an instance-level segmentation map, pix2pixHD is able to synthesize an image with a specific appearance from an instance exemplar in the same object category.
We will show that without the input instance-level pixel-wise segmentation map as a constraint, our model is still able to synthesize images with styles automatically transferred from exemplar images.  

\begin{figure*}[h]
\begin{center}
\includegraphics[width=0.9\linewidth]{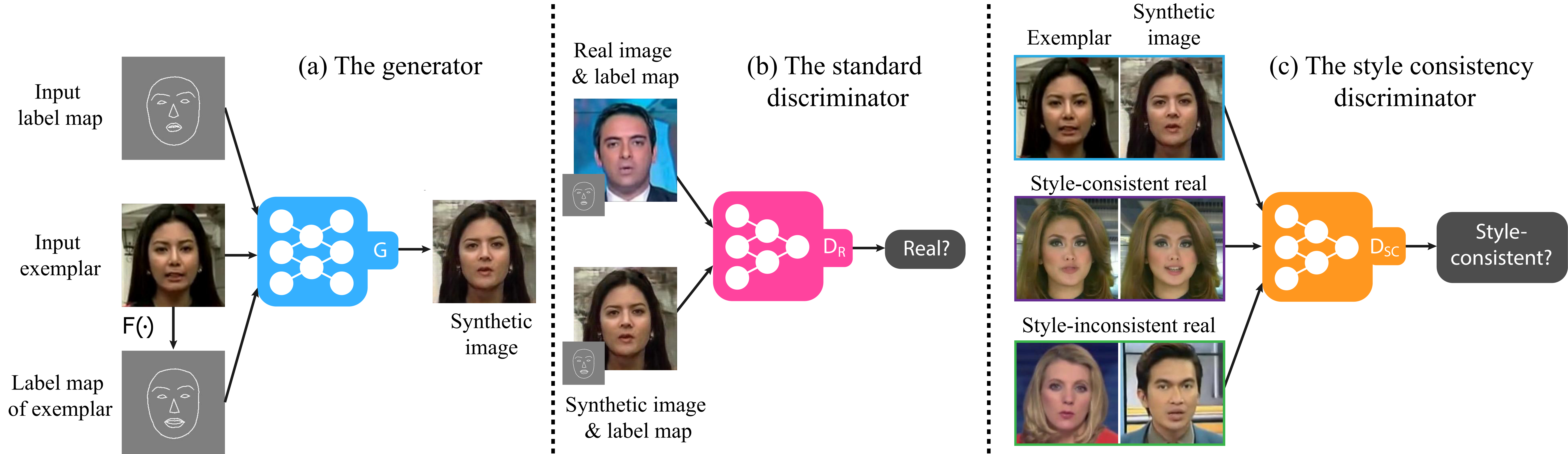}
\end{center}
\vspace{-10pt}
   \caption{Overview of our framework consisting of a generator $G$ and two discriminators $D_R$ and $D_{SC}$. (a) Given an input label map, a guided example and its labels generated by a known function $F(\cdot)$, the generator $G$ tries to synthesize an image semantically consistent to the labels, while being style-consistent to the exemplar. (b) The standard discriminator $D_R$ learns to distinguish between real and synthetic images on conditional input. (c) The style consistency discriminator $D_{SC}$ aims to distinguish between style-consistent image pairs and style-inconsistent image pairs. }
\label{fig:pipeline}
\vspace{-10pt}
\end{figure*}

\subsection{Our Model}
\label{subsec:ourmodel}
Let $I$ be a guidance image from a natural image domain $\mathcal{Y}$.
Our goal is to synthesize an image $y$, from a semantic label map $x$ and an image $I$: $(x,I) \mapsto y$. The role of $I$ is to provide a style constraint to image synthesis: the output image $y$ must be style-consistent with the exemplar $I$. Our problem is more difficult than that solved by pix2pixHD.
One particular challenge we face is that  given an input label map $x$, the ground truth images $\{y\}$ for arbitrary guided style exemplars $\{I\}$ are missing.
  To solve this weakly-supervised problem, we learn style consistency between pairs of images: they could be style-consistent image pairs $\{(y^S_1, y^S_2)\}$ or style-inconsistent image pairs $\{(y^N_1, y^N_2)\}$ (see Section \ref{sec:sample}).
  
An overview of our method is illustrated in Figure \ref{fig:pipeline}. It builds upon a single-scale version of pix2pixHD, and contains: (i) a generator $G$, with semantic map $x$, style example $I$ and its corresponding label $F(I)$  as input and output a synthetic image;  (ii) a standard discriminator $D_R$ to distinguish real images from fake ones given conditional inputs; and (iii)  we introduce a \emph{style consistency} discriminator $D_{SC}$ to detect whether the synthetic image and the guidance image $I$ are style-compatible, which operates on image pairs from domain $\mathcal{Y}$.
Here, $F(\cdot)$ is an operator which, given an image produces a set of semantic labels that represent the image (choices of $F(\cdot)$ are given in Section \ref{sec:dataset});  for convenience $F(I)$ can be visualized as an image, provided the viewer recalls that the image contains semantic labels. Our objective function contains three losses: a \emph{standard adversarial loss}; a novel \emph{adversarial style consistency loss}; and a novel \emph{adaptive semantic consistency loss}. 

\vspace{-10pt}

\subsubsection{Standard Adversarial Loss}

We apply standard adversarial losses via the standard discriminator $D_R$ as:
\vspace{-5pt}
\begin{align}
    \mathcal{L}_{\text{StdAdv}}(G,D_R) = & \mathbb{E}_{(x,y)}[\log{D_R(x,y)}]\\
    + &\mathbb{E}_{(x,I)}[\log(1-D_R(x,G(x,I,F(I))))], \nonumber
\end{align}
where the $G$ tries to synthesize images that look similar to real images from image domain $\mathcal{Y}$ regardless of specific styles, while given an image conditioned with the corresponding label map, the $D_R$ aims to determine the image is real or fake.

\vspace{-5pt}
\subsubsection{Adversarial Style Consistency Loss} \label{subsubsec:styleloss}

With the standard adversarial loss, the generator $G$ is able to synthesize images matching the data distribution of domain $\mathcal{Y}$, however the synthetic results are not guaranteed to be style-consistent with the corresponding guidance $I$. We introduce the style consistency loss $\mathcal{L}_{\text{SCAdv}}$ using a discriminator $D_{SC}$ associated with a pair of images --- either both real,  or one real  and one synthetic:
\vspace{-3pt}
\begin{equation}
\begin{split}
   & \mathcal{L}_{\text{SCAdv}}(G,D_{SC})\\
   & = \mathbb{E}_{(y^{S}_1, y^{S}_2)}[\log D_{SC}(y^S_1, y^S_2)] \\ 
   & + \mathbb{E}_{(y^{N}_1, y^{N}_2)}[\log (1-D_{SC}(y^N_1, y^N_2))]\\
   & +  \mathbb{E}_{(x,I)}[\log{(1-D_{SC}(I, G(x,I,F(I))))}],
\end{split}
\end{equation}
where $y^S_1$ and $y^S_2$ are a pair of sampled real images from domain $\mathcal{Y}$ with the same style, $y^N_1$ and $y^N_2$ are a pair of sampled real images from domain $\mathcal{Y}$ with different styles. We introduce the data sampling strategy in Section \ref{sec:sample}.

With the proposed adversarial style consistency loss $\mathcal{L}_{\text{SCAdv}}$, the discriminator $D_{SC}$ tries to learn awareness of style consistency between a pair of images, while the generator $G$ tries to fool $D_{SC}$ by generating an image with the same style to exemplar $I$.

\vspace{-10pt}
\subsubsection{Adaptive Semantic Consistency Loss} \label{subsubsec:labelloss}

The semantic consistency loss is introduced to reconstruct an image from a label map in the semantic sense of {\em e.g.} sketch. It may  appear we could use the error between the input labels $x$, and the predicted labels from the synthetic image, $G(x,I,F(I))$, for example  $||x-F(G(x,I,F(I)))||_1$ or  some variant thereof. However, different applications give distinct meanings to the semantic label maps, with the consequence that the gradient of the loss will, in general, vary between applications. This would mean selecting hyper-parameters $\lambda$ to combine losses on a per-application basis.

We avoid this problem by always computing semantic consistency losses between images: the synthetic image $G(x,I,F(I))$ and specifically an image $z$ which is a priori known to be consistent with a given semantic map $x$. Typically the image $z$ is drawn from the training dataset and we have $x = F(z)$.
A particular issue with our adopted scheme is that such losses will try to converge the network output on the image $z$, which by choice is photorealistic and is semantically consistent with $x$. Such behavior would work perfectly when $z$ and $I$ are sampled from images $\{(y^S_1, y^S_2)\}$ with the same style, but could force the output away from the desired style when $z$ and $I$ are ``style-wise'' different, \emph{i.e.} $(z,I)\in\{(y^N_1, y^N_2)\}$.

\begin{figure}[t]
\begin{center}
\includegraphics[width=0.9\linewidth]{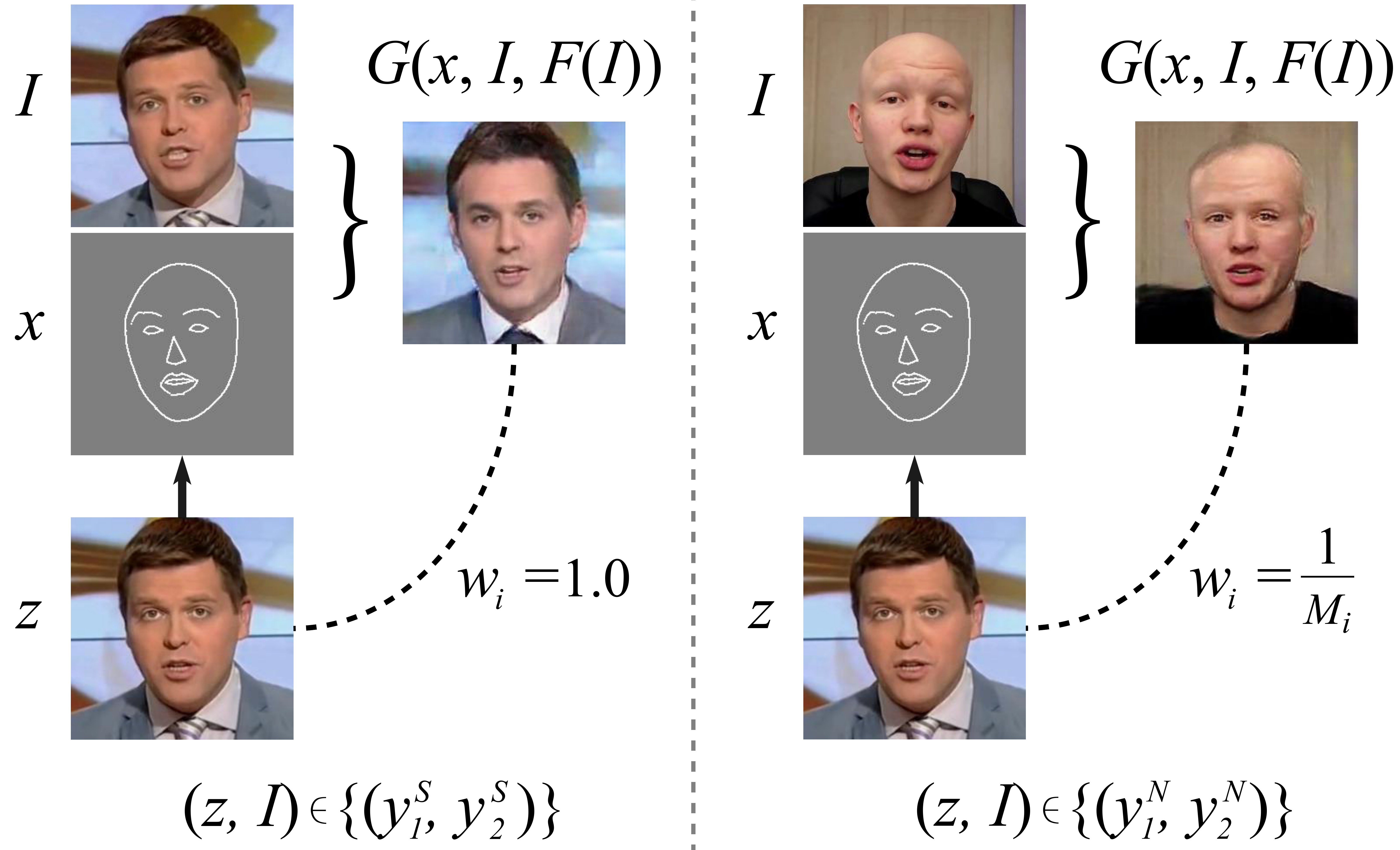}
\end{center}
\vspace{-10pt}
   \caption{Adaptive weight for semantic consistency loss. }
\label{fig:weight}
\vspace{-10pt}
\end{figure}

Our solution, is to use a novel {\em adaptive} VGG loss computed via a pre-trained model \cite{Johnson2016Perceptual} between the synthetic image $G(x,I,F(I))$ and the real image $z$ of label map $x$. An adaptive weighting scheme is proposed for per-layer VGG loss computation, to ensure the semantic consistency of the synthetic image to $x$:

\vspace{-15pt}
\begin{equation}
    \mathcal{L}_{\text{SC}}(G) = \sum_{i=1}^N w_i||L^{(i)}(z)-L^{(i)}(G(x,I,F(I)))||_1,
\end{equation}

where $L^{(i)}$ represents the $i$-th layer feature extractor of the VGG network, and $w_i$ is the adaptive weight for the $i$-th layer.  We set  $w_i=1.0$ to gain the impact of details from shallow layers when $z$ and $I$ are from style-consistent sampled pairs $\{(y^S_1, y^S_2)\}$, and $w_i=\frac{1}{M_i}$ to suppress the impact of detail matching for style-inconsistent pair $(z, I)\in \{(y^N_1, y^N_2)\}$; $M_i$ is the number of elements in the $i$-th feature layer. The adaptive weighting scheme is illustrated in Figure \ref{fig:weight}.

\vspace{-10pt}
\paragraph{Full Objective.} The final loss is formulated as:
\vspace{-5pt}
\begin{equation}
\begin{split}
  \mathcal{L}(G,D_R,D_{SC}) =  &  \mathcal{L}_{\text{StdAdv}}(G,D_R) \\
 + & \lambda_1 \mathcal{L}_{\text{SCAdv}}(G,D_{SC})\\
+ &\lambda_2 \mathcal{L}_{\text{SC}}(G),
\end{split}
\label{equ:objective}
\end{equation}
where $\lambda_1$ and $\lambda_2$ control the relative importance of the terms, our full objective is given by:
\vspace{-5pt}
\begin{equation}
    G^* = \arg\min_{G}\max_{D_R,D_{SC}} \mathcal{L}(G,D_R,D_{SC}).
\end{equation}
\vspace{-30pt}

\subsection{Sampling Strategy for Style-consistent and Style-inconsistent Image Pairs}
\label{sec:sample}
So far, we have introduced the core techniques of our network. However one prerequisite to our method is to obtain style-consistent image pairs $\{(y^S_1, y^S_2)\}$ and style-inconsistent image pairs $\{(y^N_1, y^N_2)\}$. Thus the datasets for prior image-to-image translation works \cite{pix2pix2016,pix2pixHD,CycleGAN2017,DRIT,huang2018munit} are not feasible for our training. 

A key idea for training data acquisition is to collect image pairs from videos. In face and dance synthesis tasks, we observed that: (i) within a short temporal period of a video, the style of frame contents are ensured to be the same, and (ii) frames from different videos probably have different styles (e.g. different gender, hairstyles, skin colors and make-up in the face image synthesis application). We thus randomly sample pairs of frames within $T=10$ frames from a video and regard them as style-consistent ones $\{(y^S_1, y^S_2)\}$. For style-inconsistent pairs $\{(y^N_1, y^N_2)\}$, we firstly randomly sample pairs of frames from different videos, then manually label whether images from each sampled pair are style-consistent or not.

\begin{figure}[t]
\begin{center}
\includegraphics[width=\linewidth]{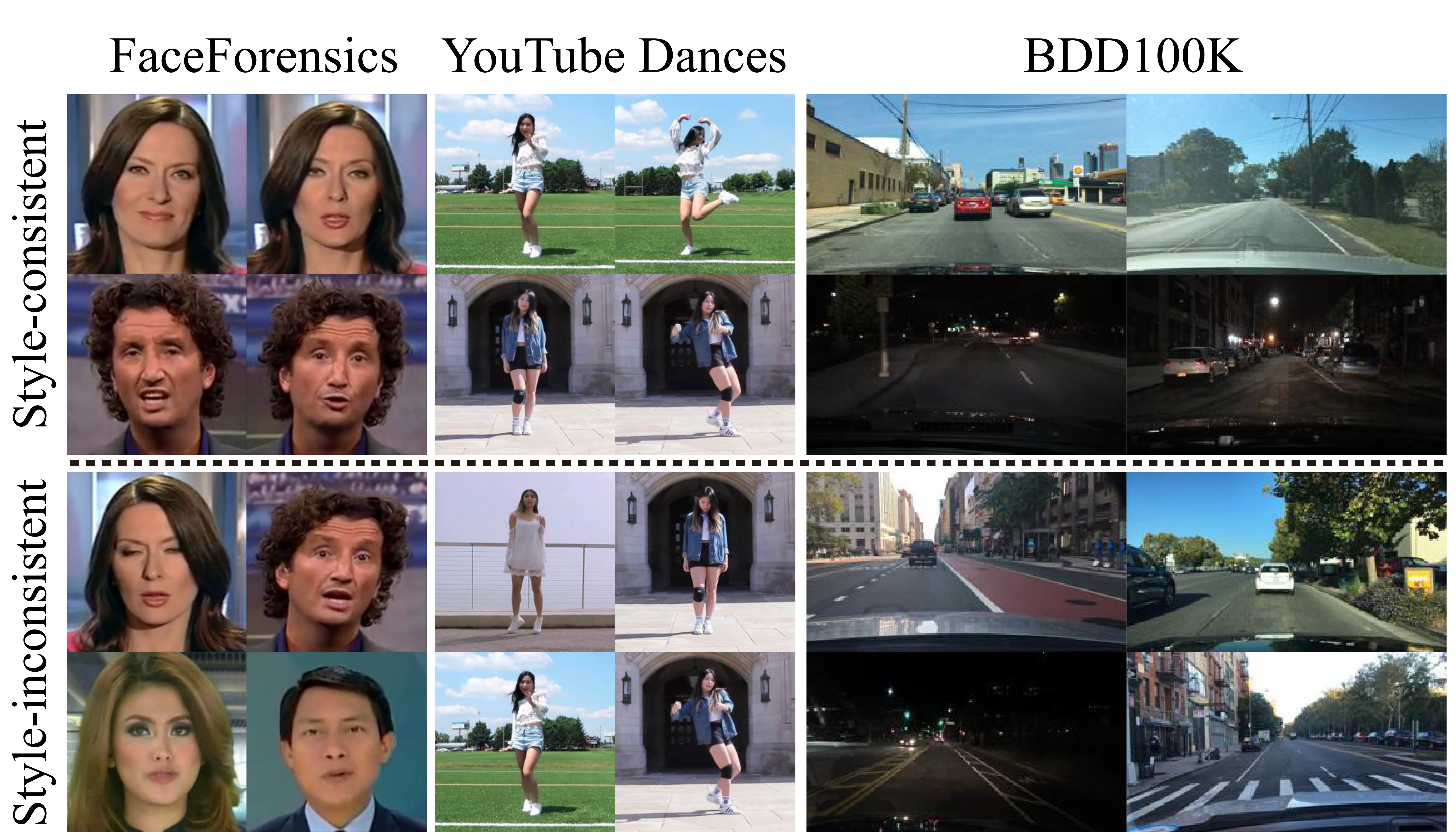}
\end{center}
\vspace{-15pt}
   \caption{Representative sampled data for training networks using FaceForensics \cite{roessler2018faceforensics}, YouTube Dances and BDD100K \cite{yu2018bdd100k} datasets. Each row shows pairs of sampled images from the above three datasets.}
\label{fig:sample}
\vspace{-10pt}
\end{figure}

In the street view synthesis task, as large scale street view videos with different styles are not easy to collect, we use images from the BDD100K dataset \cite{yu2018bdd100k}. In BDD100K, street view images and the \emph{weather}, \emph{time of day} attributes are provided. We coarsely categorize the images into $13$ style groups based on the attributes, then sample style-consistent image pairs inside each group and sample style-inconsistent image pairs between groups. 
Figure \ref{fig:sample} shows representative sampled pairs of images.

%% file: Results_Peter.tex
\section{Experiments}
\label{sec:experiments}
\subsection{Implementation Details} 
\label{sec:detail}

\begin{figure*}[t]
\begin{center}
\includegraphics[width=\linewidth]{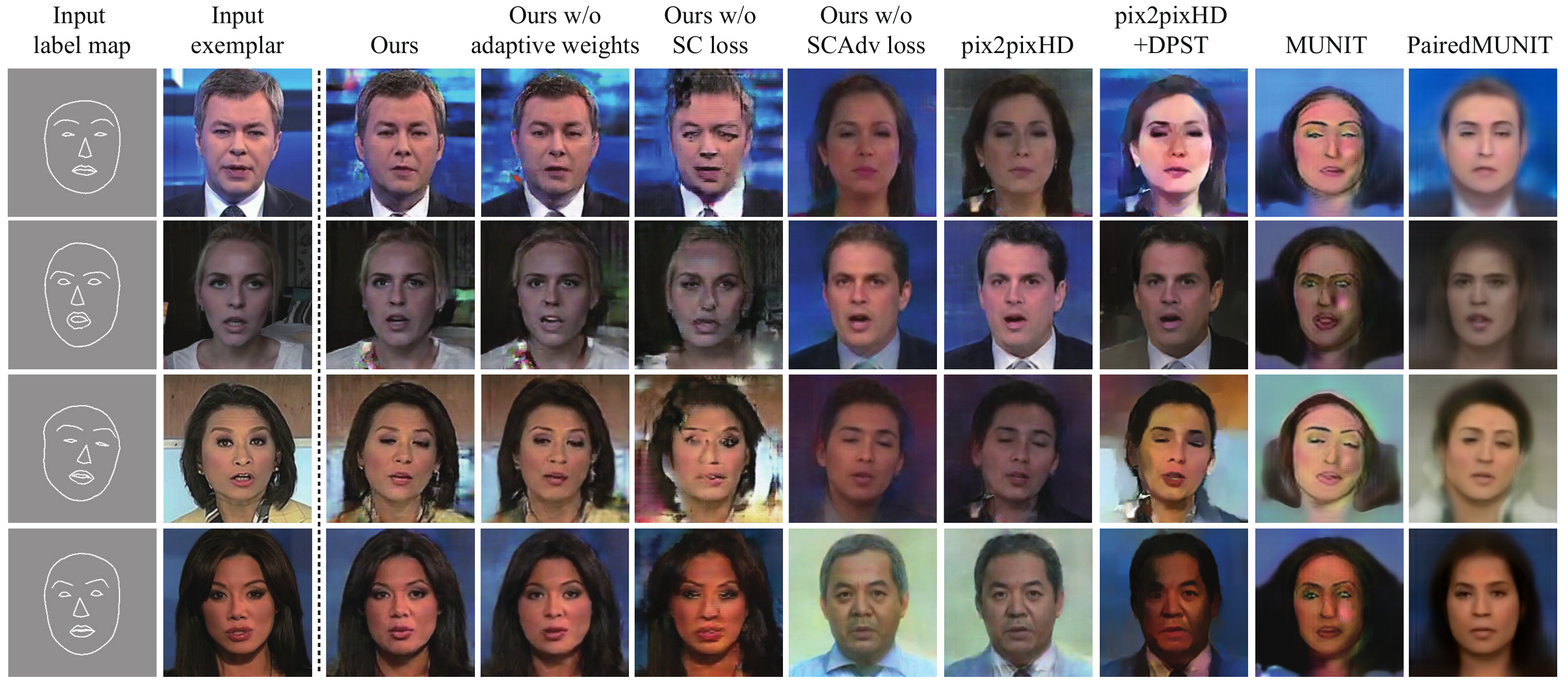}
\end{center}
\vspace{-18pt}
   \caption{Example-based face image synthesis on the FaceForensics dataset. The first column shows the input labels, the second column shows the input style example, next columns show the results from our method and our ablation studies, pix2pixHD, pix2pixHD with DPST, MUNIT and PairedMUNIT.}
\label{fig:compare_face_all}
\vspace{-10pt}
\end{figure*}
We implement our model based on the single-scale pix2pixHD framework and experiment with images with size $256\times 256$ ($256\times 144$ for street view synthesis). The generator $G$ contains several  Convolution-InstanceNorm-ReLU-Stride-2 layers to encode deep features, then 9 residual blocks \cite{he2016deep} and finally some Convolution-InstanceNorm-ReLU-Stride-0.5 layers to synthesize images. 
For both discriminators $D_R$ and $D_{SC}$, we use PatchGANs \cite{pix2pix2016} with several Convolution-InstanceNorm-LeakyReLU-Stride-2 layers with the exception that InstanceNorm is not applied in the first layer.
The slope for LeakyReLU is set as $0.2$.
For all the experiments, we set $\lambda_1 = 10$ and $\lambda_2 = 10$ in Equation \ref{equ:objective}.
All the networks are trained from scratch on an NVIDIA GTX 1080 Ti GPU using the Adam solver \cite{kingma2014adam} with a batch size of 1. The learning rate is initially fixed as $0.0002$ for the first 500K iterations and linearly decayed to zero over the next 500K iterations. We use LSGANs \cite{mao2017least} for stable training. For more details, please refer to the supplementary material.

\subsection{Datasets}
\label{sec:dataset}
We evaluate our method on face, dance and street view image synthesis tasks, using the following datasets:

\textbf{Sketch$\rightarrow$Face.} We use the real videos in the FaceForensics dataset \cite{roessler2018faceforensics}, which contains $854$ videos of reporters broadcasting news. We use the image sampling strategy described in Section~\ref{sec:sample} to acquire training image pairs from video, then apply face alignment algorithm \cite{king2009dlib} to localize facial landmarks, crop facial regions and resize them to size $256\times256$. The detected facial landmarks are connected to create face sketches as function $F(\cdot)$.

\textbf{Pose$\rightarrow$Dance.} We download $150$ solo dance videos from YouTube, crop out the central body regions and resize them to  $256\times256$. As the number of videos is small, we evenly split each video into the first part and the second part along the time-line, then sample training data only from the first parts and sample testing data only from the second parts of all the videos. The function $F(\cdot)$ is implemented using concatenated pre-trained DensePose \cite{Guler2018DensePose} and OpenPose \cite{cao2017realtime} pose detection results to provide pose labels. 

\textbf{Scene parsing$\rightarrow$Street view.}  We use the BDD100k dataset \cite{yu2018bdd100k} to synthesize street view images from pixel-wise semantic labels ({\em i.e.} scene parsing maps). We use the state-of-the-art scene parsing network DANet \cite{fu2018dual} as the function $F(\cdot)$.
Please find more details in our supplementary material.

\subsection{Baselines}
We compare our method with the following algorithms:

\textbf{pix2pixHD} and \textbf{pix2pixHD \cite{pix2pixHD} with DPST \cite{luan2017deep}.} pix2pixHD is the image-to-image translation baseline. A default image could be synthesized using pix2pixHD with its style then transfered to the guided example using Deep Photo Style Transfer (DPST) method. 

\textbf{MUNIT \cite{huang2018munit}} and \textbf{PairedMUNIT.} MUNIT is the state-of-the-art unsupervised image-to-image translation method with disentangled content and style representations that are able to translate images to given exemplars. We modify MUNIT by integrating pairwise style information to the original model and adaptively computing losses with style (denoted as PairedMUNIT).


\textbf{Ours without $\mathcal{L}_{\text{SC}}$, $\mathcal{L}_{\text{SCAdv}}$ or adaptive weights} for ablation studies. All of the methods are trained on the datasets introduced in Section \ref{sec:dataset}.

\subsection{Evaluation Metrics} 

\textbf{Photorealism and Semantic Consistency.} 
We use the Fr\'echet Inception Distance \cite{heusel2017gans} to evaluate the realism and faithfulness of the synthetic results. This metric is widely used for implicit generative models, because it correlates with the visual quality of generated samples. A smaller FID is often favored by the human subjects. We further evaluate semantic consistency by  translating the synthetic images back to the label domain and comparing the accuracy to the input labels.
For tasks Sketch$\rightarrow$Face and Pose$\rightarrow$Dance, we use the labeling endpoint error (LEPE) between the input label map $x$ and the labels generated by $F(\cdot)$ to compute the label accuracy. For task Scene parsing$\rightarrow$Street view, we use scene parsing score (SPS) \cite{fu2018dual} on synthetic street view images to measure the segmentation accuracy.

\textbf{Style Consistency.} We perform a human perceptual study to compare style consistency from human point of view. We show pairs of our result and the result from baseline methods to invited subjects and ask which one they see as being closer to the guidances' style.


\begin{table}[!t]
\centering
 \scalebox{0.85}[0.85]{
\begin{tabular}{l c c c}

\toprule[1pt]
& Sketch$\rightarrow$Face & Pose$\rightarrow$Dance & Parsing$\rightarrow$Street\\ 
\hline
pix2pixHD & 39.86 & 92.33 & 157.46\\
MUNIT &148.57 &  158.47 &  235.84\\
PairedMUNIT & 142.08&161.22 & 259.95\\
Ours& \textbf{31.26}  & \textbf{33.39} &  \textbf{96.23}\\
\bottomrule[1pt]
\end{tabular}}
\vspace{-5pt}
\caption{Photorealism comparison measured by Fr\'echet Inception Distance (FID) \cite{heusel2017gans}. }
\label{tab:is_fid}
\vspace{-5pt}
\end{table}

\begin{table}[!t]{}
\centering
 \scalebox{0.9}[0.9]{
\begin{tabular}{l c c}
\toprule[1pt]
Method & Sketch$\rightarrow$Face & Pose$\rightarrow$Dance\\
\hline
pix2pixHD & \textbf{0.0050} & \textbf{0.0163}\\
MUNIT & 0.0107 & 0.0958\\
PairedMUNIT & 0.0080& 0.0502\\
Ours & 0.0085 & 0.0186\\
\bottomrule[1pt]
\end{tabular}}
\vspace{-5pt}
\caption{Semantic consistency measured by normalized label endpoint error for different methods in face and dance image synthesis tasks.}
\label{tab:epe}
\vspace{-15pt}
\end{table}

\subsection{Results}

\textbf{Main Results.} In Figure \ref{fig:compare_face_all}, we show our results (column 3) and the results from baseline methods in the Sketch$\rightarrow$Face synthesis application on the test set. 
While the pix2pixHD is able to generate photorealistic images consistent with the input semantic labels, it is not able to keep the style (\emph{e.g.} gender, hair, skin color) from input exemplars in the synthetic results, even enhanced by the deep photo style transfer effect (column 7 and 8). The unsupervised method MUNIT and its improvement PairedMUNIT fail to generate photorealistic results from semantic maps in this application (column 9 and 10). The possible reason for their failures is that they assume that the input and output domains share the same content space, which is not true in image synthesis applications from semantic label maps. 

Table \ref{tab:is_fid} gives the quantitative evaluation of the photorealism measured by FID in various image synthesis tasks, where our method performs the best. 
The semantic consistency of synthetic results to the input labels is given by LEPE in Table \ref{tab:epe}. It can be seen that the pix2pixHD obtains the best semantic consistency to the input labels, because it does not lose semantic accuracy by totally ignoring style consistency. Our method outperforms MUNIT and PairedMUNIT.

For style consistency evaluation, we conduct a human perception study commonly used in image-to-image translation works \cite{pix2pix2016,CycleGAN2017,chen2017photographic,pix2pixHD,StarGAN2018}. The input exemplars and pairwise synthetic results sampled from our method and a baseline method are shown to the subjects with unlimited watching time. Then the subjects were asked ``Which image is closer to the exemplar in terms of style?" Images for user study were randomly sampled from the test set; each pair was shown in random order and guaranteed to be examined by at least 30 subjects. The ratios of votes our method got over baseline methods are given in Table \ref{tab:human}. Our method won more user preferences in pairwise comparison. 
The quantitative results shown that our results are more photorealistic and more style-consistent with the exemplars.

We conducted ablation studies to verify our model. As can be seen in Figure \ref{fig:compare_face_all}, without the adaptive weight scheme in $\mathcal{L}_{\text{SC}}$, the quality of results is slightly reduced;  without the semantic loss $\mathcal{L}_{\text{SC}}$, the semantic consistency would lose; without the style consistency adversarial loss $\mathcal{L}_{\text{SCAdv}}$, the target style is not maintained. Quantitative photorealism statistics reported in Table \ref{tab:ablation} validated the above observation. We further extract $50\times50$ eye patches from synthetic images and exemplars and compute the VGG feature distance between them. Table \ref{tab:VGG} indicates that the weight adaptation makes a quantitative improvement of style consistency. 

\begin{figure}[!t]
\begin{center}
\includegraphics[width=\linewidth]{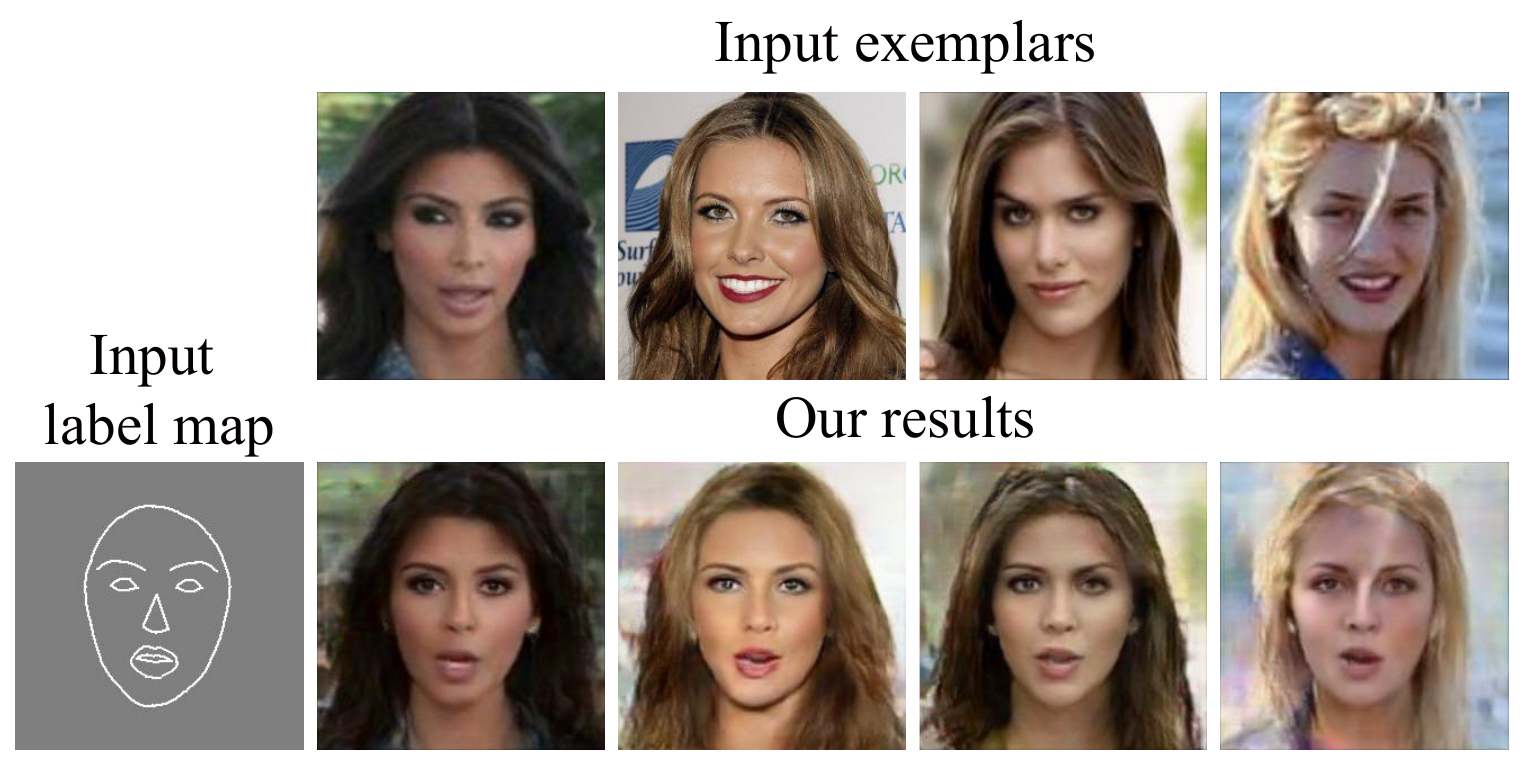}
\end{center}
\vspace{-18pt}
   \caption{In-the-wild Sketch$\rightarrow$Face synthesis.}
\label{fig:inthewild}
\vspace{-8pt}
\end{figure}

Figure \ref{fig:inthewild} shows the in-the-wild synthesis results from our model using Internet images. The results indicate that the model generalizes well for ``unseen'' cases. We provide more results in the supplementary material.

\begin{table}[!t]
\centering{}
 \scalebox{0.9}[0.9]{

{\begin{tabular}{c c c c}
\toprule[1pt]
 & pix2pixHD & pix2pixHD+DPST & PairedMUNIT \\
\hline
Ours & 89.12\% & 80.67\% & 90.88\%\\
\bottomrule[1pt]
\end{tabular}}}
\vspace{-5pt}
\caption{Style consistency evaluation by human option study on Sketch$\rightarrow$Face synthesis. Each cell lists the percentage where our result is preferred over the other method.}
\label{tab:human}
\vspace{-15pt}
\end{table}




\begin{table}[!t]
\centering

 \scalebox{0.88}[0.88]{
\begin{tabular}{ c c | c c}
\toprule[1pt]
 Method & FID & Method & FID\\
\hline
Ours w/o adaptive weights & 35.59 & Ours w/o $\mathcal{L}_{\text{SCAdv}}$ &  58.08 \\
Ours w/o $\mathcal{L}_{\text{SC}}$ & 76.59 & Ours  &  \textbf{31.26}\\
\bottomrule[1pt]
\end{tabular}
}
\vspace{-5pt}
\caption{Ablation study: Fr\'echet Inception Distance (FID) of our results and alternatives on the Sketch$\rightarrow$Face synthesis task.}
\label{tab:ablation}
\vspace{-5pt}
\end{table}

\begin{table}[!t]
\centering
 \vspace{-5pt}
 \scalebox{0.78}[0.78]{

\begin{tabular}{c c c c c}
\toprule[1pt]
 & Ours & Ours w/o adapt. weights& w/o $\mathcal{L}_{\text{SC}}$ & w/o $\mathcal{L}_{\text{SCAdv}}$ \\
\hline
VGG Dist. &\textbf{0.643} & 0.654& 0.898 & 0.703 \\
\bottomrule[1pt]
\end{tabular}
}
\vspace{-8pt}
\caption{VGG feature distance of eye patches between synthetic image and exemplar.}
\label{tab:VGG}
\vspace{-12pt}
\end{table}

\begin{figure}[!t]
\begin{center}
\includegraphics[width=0.96\linewidth]{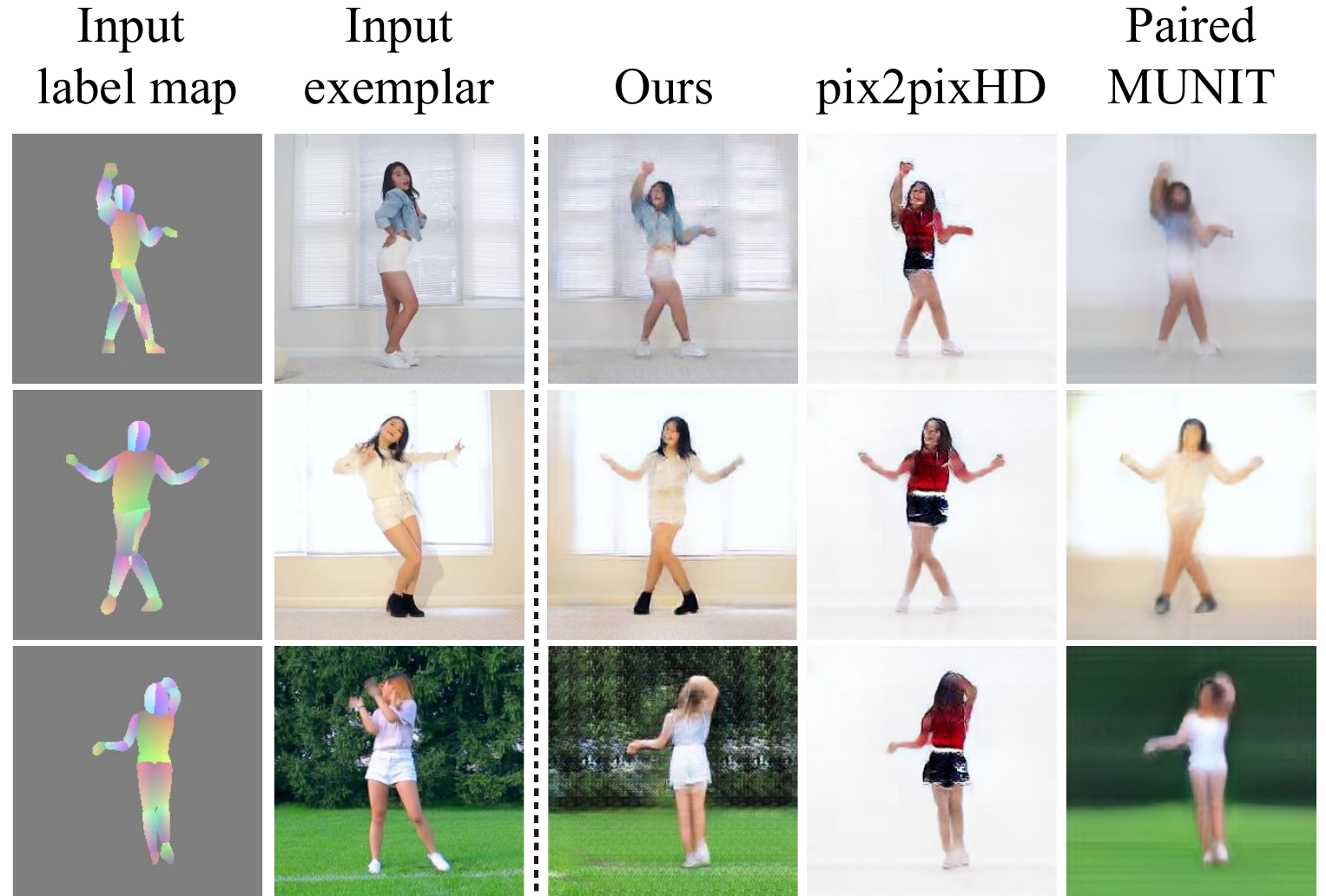}
\end{center}
\vspace{-15pt}
   \caption{Dance synthesis from pose maps.}
\label{fig:dance_compare}
\vspace{-8pt}
\end{figure}

\begin{figure}[!t]
\begin{center}
\includegraphics[width=0.8\linewidth]{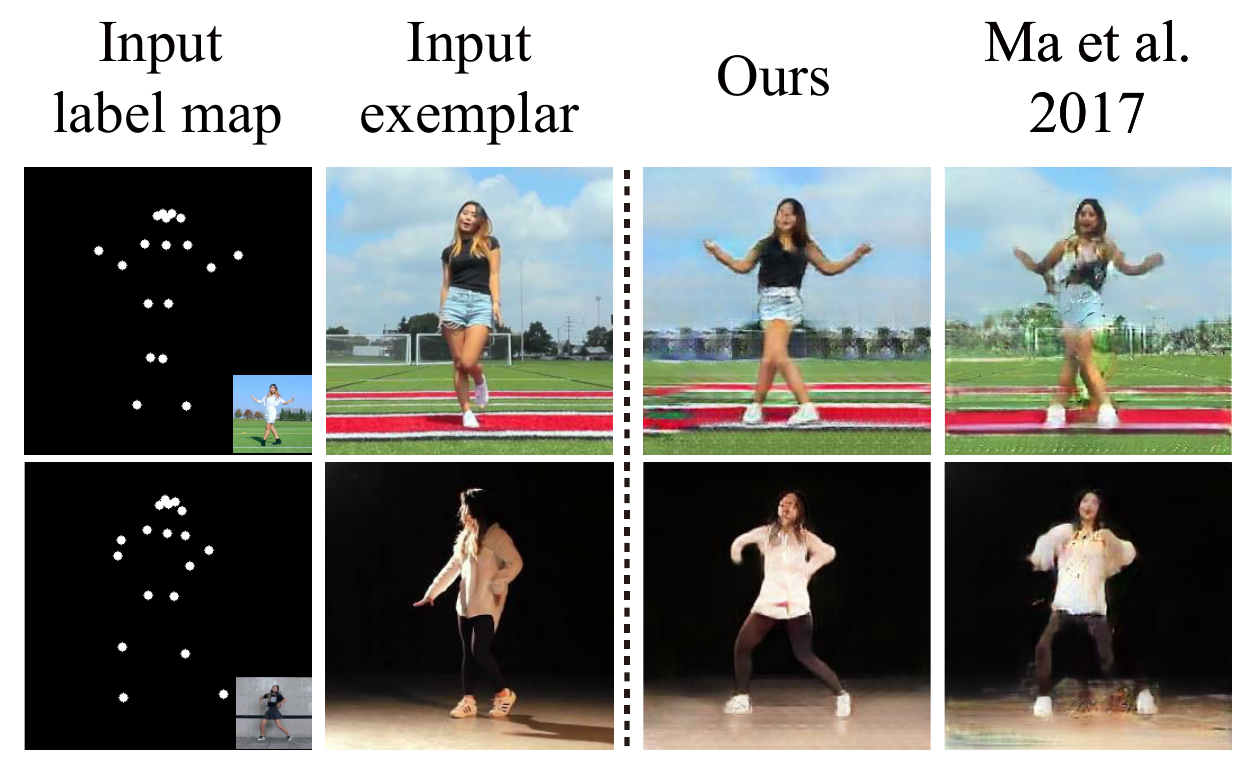}
\end{center}
\vspace{-15pt}
   \caption{Pose$\rightarrow$Dance comparison with Ma \etal \cite{DBLP:conf/nips/MaJSSTG17}.}
\label{fig:compare_ma}
\vspace{-15pt}
\end{figure}

\begin{figure*}[t]
\begin{center}
\includegraphics[width=\linewidth]{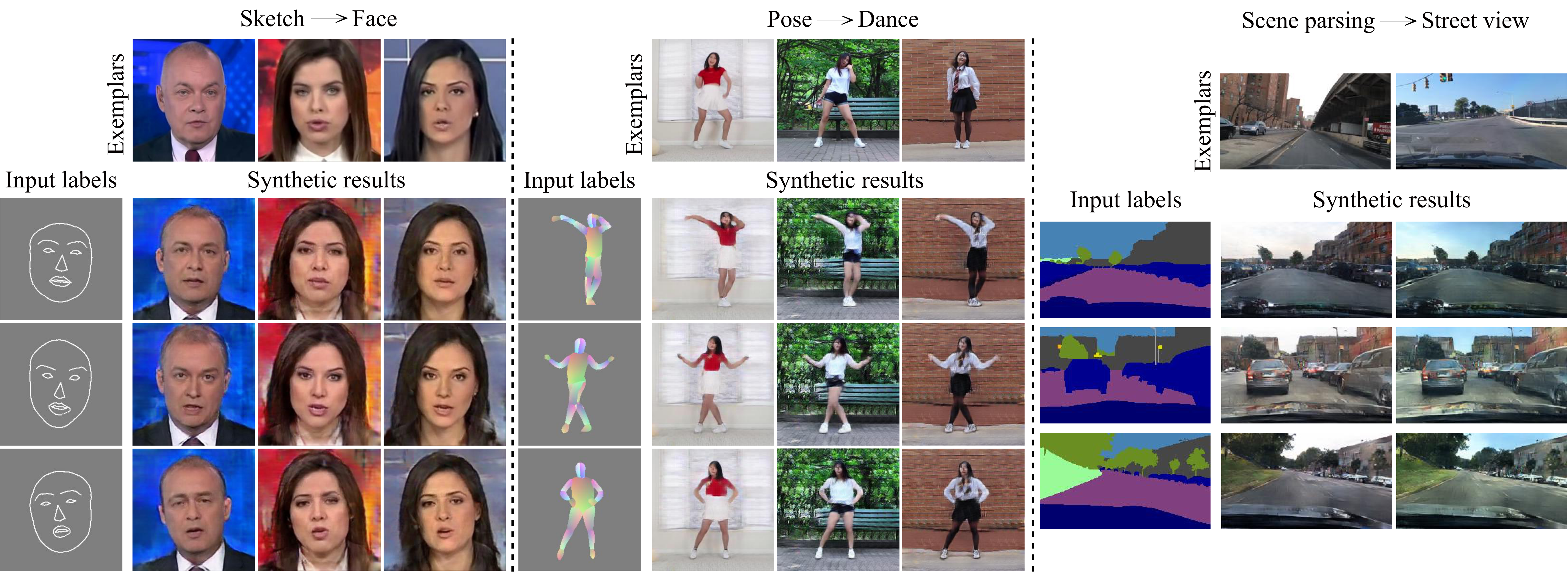}
\end{center}
\vspace{-15pt}{}
   \caption{More results of example-based image synthesis on face, dance and street view synthesis tasks.}
\label{fig:all}
\vspace{-10pt}
\end{figure*}

\begin{figure}[t]
\begin{center}
\includegraphics[width=\linewidth]{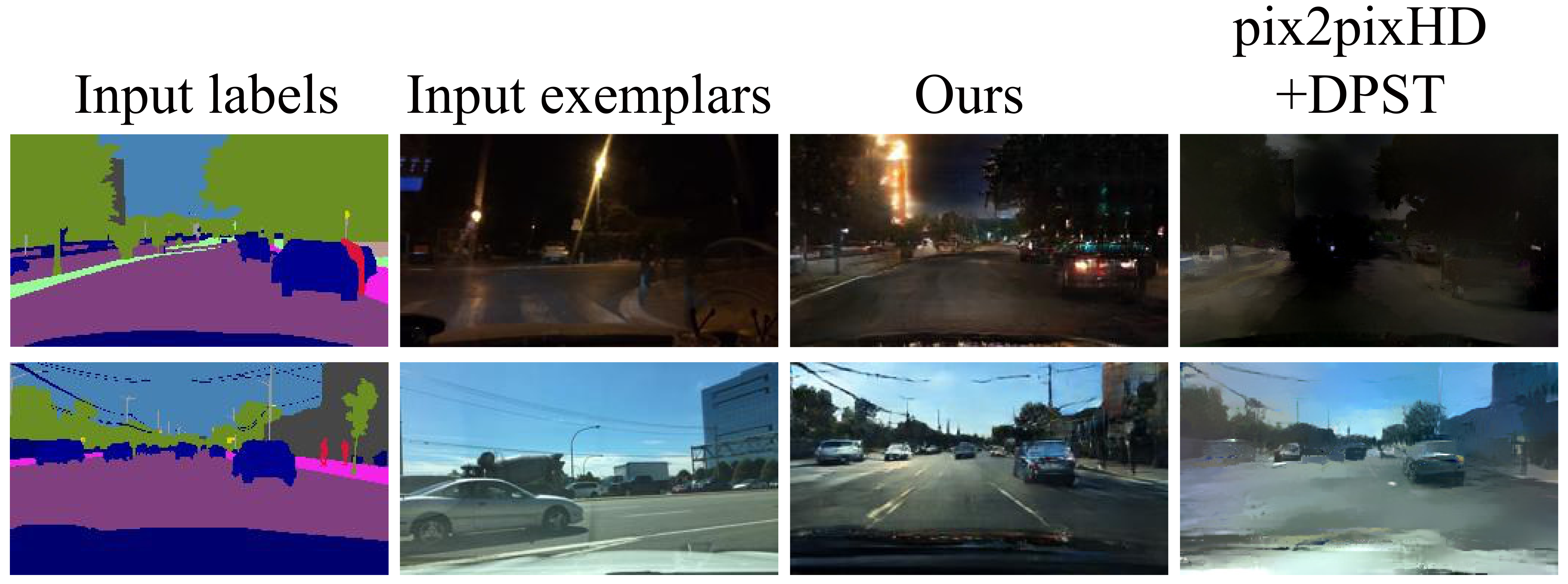}
\end{center}
\vspace{-18pt}
   \caption{Street view synthesis from scene parsing maps and corresponding exemplars.}
\label{fig:street_compare}

\end{figure}


\textbf{Pose$\rightarrow$Dance Synthesis.} Figure \ref{fig:dance_compare} shows a visual comparison of our method and baselines in the Pose$\rightarrow$Dance synthesis application. The semantic consistency of synthetic results to the input labels measured using LEPE are given in Table \ref{tab:epe}. Although the facial regions of our results are blurry without including facial landmarks in the input pose labels, our model still produces images that are style-consistent with the guidance images while consistent with the semantic labels. Figure \ref{fig:compare_ma} shows the visual comparison with Ma \etal \cite{DBLP:conf/nips/MaJSSTG17} on the dancing dataset. The generated poses and clothes in our results are visually better.

\textbf{Scene parsing$\rightarrow$Street view Synthesis.} A comparison of our method and baselines in the Scene parsing$\rightarrow$Street view task is given in Figure \ref{fig:street_compare}.   The semantic consistency of synthetic results to the input labels measured using SPS are given in Table \ref{tab:fcnscore}.  Although the scene in the guidance images are not quite the same as the semantics of the input label maps, our model is able to produce images that are semantically consistent with the segmentation map and style-consistent with the guidance image.




Figure \ref{fig:all} shows more results. Our network can faithfully synthesize images from various semantic labels and exemplars. Please find more results in the supplementary file.

\begin{table}[!t]
\centering
 \scalebox{0.9}[0.9]{

{\begin{tabular}{l c c c}
\toprule[1pt]
Method & Per-pixel acc. & Per-class acc. & Class IOU\\
\hline
pix2pixHD & 83.85 & 36.17 & 0.310\\
MUNIT & 58.58 & 18.99 &0.139\\
PairedMUNIT & 62.96 & 22.41 &0.160\\
Ours & \textbf{84.71} & \textbf{39.44} & \textbf{0.333}\\
\hline
Original image & 86.74 & 52.25 & 0.452\\
\bottomrule[1pt]
\end{tabular}}}
\vspace{-5pt}
\caption{Semantic consistency measured by scene parsing score \cite{fu2018dual} for different methods on the street view image synthesis task.}
\label{tab:fcnscore}
\vspace{-15pt}
\end{table}